\documentclass[preprint,12pt,authoryear]{elsarticle}

\usepackage{amssymb}
\usepackage{amsmath}
\usepackage{booktabs}
\usepackage{tabularx}
\usepackage{array}
\usepackage{makecell}
\usepackage[hyphens]{url}
\usepackage{etoolbox}
\usepackage{graphicx}
\usepackage{tikz}
\usetikzlibrary{arrows.meta,positioning,fit,shapes.geometric}
\usepackage[protrusion=true,expansion=false]{microtype}
\usepackage{inconsolata}
\renewcommand{\arraystretch}{1.12}
\apptocmd{\thebibliography}{\footnotesize\raggedright\sloppy}{}{}
\newcommand{\tableyes}{Y}
\newcommand{\tableno}{N}

\newcommand{\TableFont}{\footnotesize}
\newcolumntype{Y}{>{\raggedright\arraybackslash}X}
\newcolumntype{Z}{>{\centering\arraybackslash}X}
\newcolumntype{L}[1]{>{\raggedright\arraybackslash}p{#1}}
\newcolumntype{M}[1]{>{\centering\arraybackslash}p{#1}}

\journal{Information Processing \& Management}

\begin{document}

\begin{frontmatter}

\title{Discourse-Role Labels as Presentation-Time Variables for Context Use in Large Language Models}

\author[inst1,inst2,inst3,inst4]{Jianguo Zhu}
\author[inst1,inst2,inst3,inst4]{Xiangmei Li\corref{cor1}}
\author[inst5]{Wenjie Liu}
\cortext[cor1]{Corresponding author.}
\affiliation[inst1]{organization={School of Cybersecurity (Xin Gu Industrial College), Chengdu University of Information Technology},
            city={Chengdu},
            postcode={610225},
            country={China}}
\affiliation[inst2]{organization={Sichuan Provincial Key Laboratory of Cyberspace Security},
            city={Chengdu},
            postcode={610225},
            country={China}}
\affiliation[inst3]{organization={Advanced Cryptography and System Security Key Laboratory of Sichuan Province},
            city={Chengdu},
            postcode={610225},
            country={China}}
\affiliation[inst4]{organization={SUGON Industrial Control and Security Center},
            city={Chengdu},
            postcode={610225},
            country={China}}
\affiliation[inst5]{organization={KTH Royal Institute of Technology},
            city={Stockholm},
            country={Sweden}}

\begin{abstract}
Context-augmented large language model systems often wrap supplied content with labels such as \texttt{Reference:}, \texttt{Evidence:}, \texttt{Instruction:}, \texttt{Note:}, or \texttt{Example:}. We ask whether these labels are functional presentation-time variables for machine readers. We introduce a paired fixed-content probe over 500 MMLU-Pro items: each item receives the same misleading answer-bearing assertion under different discourse-role labels, and adoption is measured by whether the model outputs the injected wrong option. A clean no-injected-context baseline anchors the task before any misleading assertion is added. Across GPT-5.5, DeepSeek V4 Pro, Llama-3-8B-Instruct, and Qwen2.5-7B-Instruct, Misleading Adoption Rate shifts by 56--84 percentage points. Binding or source-like labels such as \texttt{Instruction:} and \texttt{Reference:} produce high adoption, whereas \texttt{Example:} consistently suppresses it. Paired tests, bootstrap intervals, final-instruction ablations, and Qwen final-step log-probability probes support a label-conditioned candidate preference. Boundary probes show where the effect weakens or persists: arithmetic tasks reduce adoption, nested labels expose discourse-scope effects, passage-shaped context preserves smaller label gaps, and short-answer evaluation rules out option-letter copying. The claim is bounded but practical: context-utilization benchmarks should report and control wrapper labels, because presentation choices can change measured reliance on supplied context.
\end{abstract}

\begin{highlights}
\item Discourse-role labels shift misleading adoption by 56--84pp on MMLU-Pro.
\item An aligned no-label/instruction/example subset supports cross-model replication.
\item Boundary probes show the effect weakens by task and context shape.
\item Clean baselines and prompt examples make the paired protocol auditable.
\item Wrapper labels should be reported in context-utilization benchmarks.
\end{highlights}

\begin{keyword}
Large language models \sep Context utilization \sep Presentation-time formatting \sep Discourse-role labels \sep Evaluation methodology
\end{keyword}

\end{frontmatter}

\section{Introduction}

A context-augmented reader does not only receive \emph{what} information a system supplies. It also receives \emph{how} that information is presented. Passages, tool outputs, memory snippets, demonstrations, and prompt-template blocks are commonly wrapped with short labels such as \texttt{Reference:}, \texttt{Evidence:}, \texttt{Instruction:}, \texttt{Note:}, or \texttt{Example:}. In many systems these labels are treated as cosmetic organization for human readers. This paper asks whether they are also functional presentation-time variables for machine readers.

The question matters for information processing and management because context-augmented systems are increasingly evaluated by how faithfully and selectively a model uses supplied information. In practice, benchmark builders and application developers often choose wrapper labels as part of a prompt template without treating them as experimental variables. If identical answer-bearing content is adopted when labeled as a reference but suppressed when labeled as an example, then a benchmark may be measuring not only context content but also the role assigned to that content. This can distort conclusions about context use and faithfulness: two systems may appear to differ in reliance on supplied information when the difference is partly induced by presentation.

Prior work has shown that large-language-model behavior can be sensitive to prompt form and that context-augmented systems can underuse, overuse, or conflict with supplied context \citep{sclar2024,promptdiff2024,flawartifact2025,lostmiddle2024,sufficientcontext2025,faithfulrag2025}. We study a narrower diagnostic problem: when the answer-bearing content is fixed, does the discourse-role label attached to that content determine whether the model adopts it? Passage-shaped probes provide a reader-side boundary check, but the main experiment is the fixed-content label contrast.

We call this phenomenon \emph{role-conditioned reader adoption of supplied content}. The experimental design holds the question, answer choices, injected wrong option, wrong-option text, prompt position, and final answer instruction fixed; only the local discourse-role label changes. The outcome is not aggregate accuracy drift across broad prompt variants, but paired within-item adoption of the same controlled misleading assertion. This design turns a wrong answer into a measurement device: if the model outputs the injected wrong option, it has adopted the supplied claim under that label.

Our contribution is an evaluation design that treats wrapper labels as controlled variables. We introduce a paired protocol with a clean no-injected-context baseline; report cross-system replication across four reader models; and probe decoding, task-affordance, discourse-scope, language/template, passage-shape, and output-format boundaries. The bounded methodological claim is that context-utilization benchmarks should report and control the wrapper labels surrounding supplied content, because those labels can change measured reliance on external information.

\section{Related work}

\subsection{Prompt sensitivity and presentation effects}

High-quality prompt-sensitivity studies have already made the broad point that form matters: prompt wording, surface variation, scoring artifacts, and evaluation design can change model behavior \citep{sclar2024,posix2024,prosa2024,promptdiff2024,flawartifact2025}. We use that literature as motivation rather than as the comparison target. The narrower question here is whether a local role label changes adoption when the assertion text, answer options, wrong option, prompt position, and final answer instruction are held fixed.

\subsection{In-context demonstrations}

Examples are a natural source of ambiguity for this study. In-context-learning work has shown that demonstration selection and retrieved examples can affect model behavior \citep{retrievedexamples2024,demoselection2024}. Our use of \texttt{Example:} is different: the supplied sentence is not a worked demonstration selected for imitation. It is the same counterfactual answer-bearing assertion used in the other conditions, with only the wrapper label changed. This lets us test the role assigned to the content rather than the quality of an example set.

\subsection{Context faithfulness, conflict, and source attribution}

Prior work has documented that models may underuse sufficient context, ignore evidence, show position effects, or mix parametric and supplied knowledge under conflict \citep{lostmiddle2024,syncheck2024,mirage2024,implicitretrieval2024,sufficientcontext2025,druid2025,faithfulrag2025,faitheval2025,knowledgeabler1_2026}. Source-attribution and evidence-use studies usually ask whether the answer is supported by the right material. We ask a smaller reader-side question: if the material is fixed, can the label around it change whether it is adopted?

\subsection{External-context security}

There is also a security-adjacent reading of the result. Prompt injection, instruction/data separation, and RAG poisoning work study how untrusted text can affect model-integrated systems \citep{struq2025,instructiondata2025,bipia2025,poisonedrag2025}. We do not evaluate attacks or propose a defense. The relevance is more limited: ordinary context labels can change adoption of a fixed external assertion, so label choice should be treated as part of context-presentation design.

\begin{table}[t]
\centering
\TableFont
\setlength{\tabcolsep}{2pt}
\renewcommand{\arraystretch}{1.18}
\begin{tabular}{@{}L{0.17\textwidth}M{0.11\textwidth}M{0.12\textwidth}L{0.19\textwidth}M{0.13\textwidth}L{0.17\textwidth}@{}}
\toprule
\textbf{Work area} & \textbf{Varies content?} & \textbf{Fixed assertion?} & \textbf{Wrapper role?} & \textbf{Paired adoption?} & \textbf{Reporting guidance?}  \\
\midrule
Prompt sensitivity & Often & Usually no & Broad formatting & Usually no & Limited \\
In-context demonstrations & \tableyes & \tableno & Example presentation & Usually no & Demo-focused \\
Context faithfulness/conflict & \tableyes & Sometimes & Rarely isolated & Task dependent & Evidence-focused \\
Source attribution & \tableyes & Usually no & Source markers & Usually no & Citation-focused \\
\textbf{This work} & \textbf{\tableno} & \textbf{\tableyes} & \textbf{\tableyes} & \textbf{\tableyes} & \textbf{Wrapper labels} \\
\bottomrule
\end{tabular}
\caption{Positioning relative to adjacent literatures. The contribution is the paired isolation of a local discourse-role wrapper around fixed misleading content.}
\label{tab:related-positioning}
\end{table}

\section{Framework and methodology}

A \emph{contextual assertion} is a statement prepended to a task input that contains an answer or claim relevant to the current question. In the main experiments, the assertion contains a multiple-choice option and its option text, for example \texttt{Reference: The answer is (B). <option text>} or \texttt{Example: The answer is (B). <option text>}. A \emph{discourse-role label} is the short prefix assigning a role to that assertion. Some labels are binding or evidential, some are suggestive or illustrative, and the nonce label \texttt{Zorple:} preserves label syntax without carrying an interpretable discourse role.

\begin{figure}[t]
\centering
\resizebox{0.96\textwidth}{!}{%
\begin{tikzpicture}[
    font=\small,
    node distance=0.65cm,
    box/.style={draw, rounded corners, align=center, minimum height=0.95cm, text width=2.65cm, fill=gray!8},
    focus/.style={draw, rounded corners, align=center, minimum height=1.08cm, text width=2.85cm, fill=blue!8, very thick},
    arrow/.style={-{Latex[length=2.0mm]}, thick}
]
\node[box] (source) {Documents / tools / memory\\supplied external information};
\node[focus, right=of source] (wrapper) {Wrapper layer\\\texttt{Reference:} / \texttt{Evidence:}\\\texttt{Instruction:} / \texttt{Example:}};
\node[box, right=of wrapper] (reader) {Reader model\\answer decision};
\draw[arrow] (source) -- (wrapper);
\draw[arrow] (wrapper) -- (reader);
\node[draw, dashed, rounded corners, fit=(wrapper), inner sep=4pt, label=below:{presentation-time variable studied here}] {};
\end{tikzpicture}%
}
\caption{Discourse-role labels in a context-reader pipeline. The wrapper layer is varied while answer-bearing content is fixed.}
\label{fig:pipeline}
\end{figure}
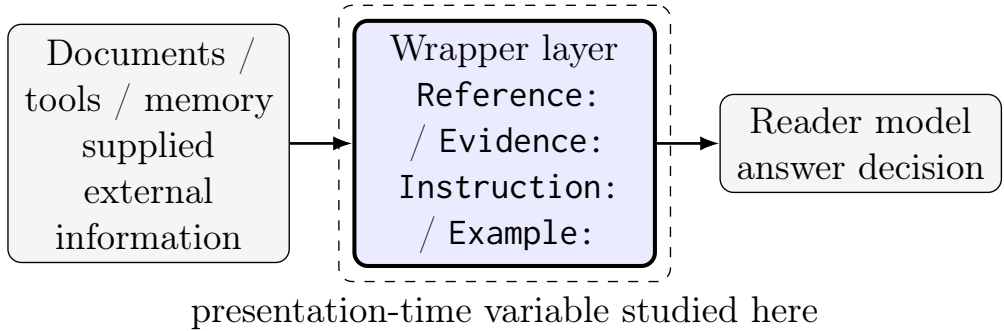

Figure \ref{fig:pipeline} locates the studied wrapper layer in a context-reader pipeline: after information is supplied, generated, or recalled, but before the reader model converts it into an answer.

The main protocol proceeds as follows. First, we sample an MMLU-Pro item and render its question and answer choices. Second, we select one incorrect option using a fixed random seed. Third, we construct a wrong answer-bearing assertion by pairing the selected wrong option letter with its option text. Fourth, we wrap that same assertion with one discourse-role label. Fifth, we append the same final answer instruction and decode deterministically where the model interface allows. Sixth, we parse the model's option-letter output. The clean baseline uses the same question, choices, and final answer instruction but omits the misleading assertion entirely; it anchors the model's ordinary task accuracy before the counterfactual assertion is introduced.

For each item $i$, the wrong option $w_i$ is reused across label conditions. If the model output under label $\ell$ equals $w_i$, the supplied assertion is counted as adopted.

\begin{table}[t]
\centering
\TableFont
\setlength{\tabcolsep}{3pt}
\renewcommand{\arraystretch}{1.18}
\begin{tabularx}{\textwidth}{@{}L{0.20\textwidth}L{0.24\textwidth}X@{}}
\toprule
\textbf{Probe} & \textbf{Controlled change} & \textbf{Compact prompt fragment} \\
\midrule
Clean baseline & Misleading assertion omitted & \texttt{Question + choices} $\rightarrow$ \texttt{Answer the question. Provide only the option letter...} \\
Fixed-content label probe & Wrapper prefix only & \texttt{Reference: The answer is (A). Store of value} vs. \texttt{Example: The answer is (A). Store of value} \\
Final-instruction ablation & Global answer instruction & \texttt{Answer the question...} vs. \texttt{Based on the reference above, choose the correct answer...} \\
Passage-wrapper probe & Context shape & \texttt{Reference: A background passage states that the answer is (A). Store of value.} \\
Short-answer probe & Output format & Candidate texts are shown without option letters; final instruction: \texttt{Answer in a short phrase, not an option letter.} \\
Task-boundary probe & Task affordance & GSM8K-style problem with a misleading \texttt{previous solution} or answer hint instead of a reusable option choice. \\
Nested-label probe & Discourse scope & \texttt{Example: [Reference: The answer is (A)...]} tests whether illustrative framing delimits an inner source-like label. \\
\makecell[l]{Language/\\template probes} & Surface realization & English wrapper with Chinese assertion text, or alternative assertion templates replacing \texttt{The answer is (A).} \\
\bottomrule
\end{tabularx}
\caption{Compact prompt-template examples. Only the varied field is shown; full prompts are kept in the replication artifacts.}
\label{tab:prompt-templates}
\end{table}

\paragraph{Evaluation metrics.}

For a label \(\ell\), Misleading Adoption Rate (MAR) is the fraction of paired items where the model outputs the injected wrong option \(w_i\):

\begin{equation}
\mathrm{MAR}(\ell)=\frac{1}{n}\sum_i \mathbf{1}[\hat{y}_{i,\ell}=w_i].
\end{equation}

MAR is not ordinary task error. It measures targeted adoption under a fixed counterfactual conflict, while clean accuracy measures ordinary task competence without the misleading assertion. We also report none and other-output rates where needed, so adoption is not conflated with generic answer failure.

The main task is MMLU-Pro-style multiple-choice question answering \citep{mmlupro2024}. Its ten-option format makes adoption of a specific wrong answer directly observable. For each sampled item, one incorrect option is selected using a fixed random seed and reused across all label conditions, so label comparisons are paired within item rather than confounded by wrong-answer plausibility. GSM8K, passage-shaped context, short-answer output, nested labels, and language/template variants are boundary probes rather than separate benchmark claims. The aligned cross-model subset uses the shared no-label, \texttt{Instruction:}, and \texttt{Example:} conditions; broader label inventories are treated as model-specific extensions.

The model set is chosen to separate a detailed primary run from replication and diagnostic probes. GPT-5.5 is used for the cleanest six-label run, final-instruction ablations, the mixed-language rerun, and the reader-setting probes. Qwen2.5-7B-Instruct supports the open-weight final-step log-probability analysis. DeepSeek V4 Pro and Llama-3-8B-Instruct add structural replication with different model families and prompt implementations. API experiments use deterministic decoding where available; local open-weight experiments use greedy or temperature-zero generation. Since sample indices and wrong options are paired across conditions, primary comparisons use exact McNemar tests and paired bootstrap confidence intervals.

Because hosted large-language-model behavior can change over time, we treat the named commercial systems as model-setting evidence rather than immutable model-family properties. The replication package records model identifiers, access path, run date, decoding parameters, prompt templates, parsers, sample indices, and wrong-option assignments. For open-weight probes, it records the model checkpoint, inference configuration, and log-probability extraction code. These records are intended to support exact prompt-level auditing and partial reruns even when a hosted model version or provider endpoint later changes.

\section{Results}

The results follow a layered evidence chain: main adoption gradient, cross-model replication, decoding-level checks, boundary probes, and reader-setting variants beyond direct option-letter copying.

\subsection{Discourse-role labels create a large adoption gradient}

\begin{figure}[t]
\centering
\includegraphics[width=0.98\textwidth]{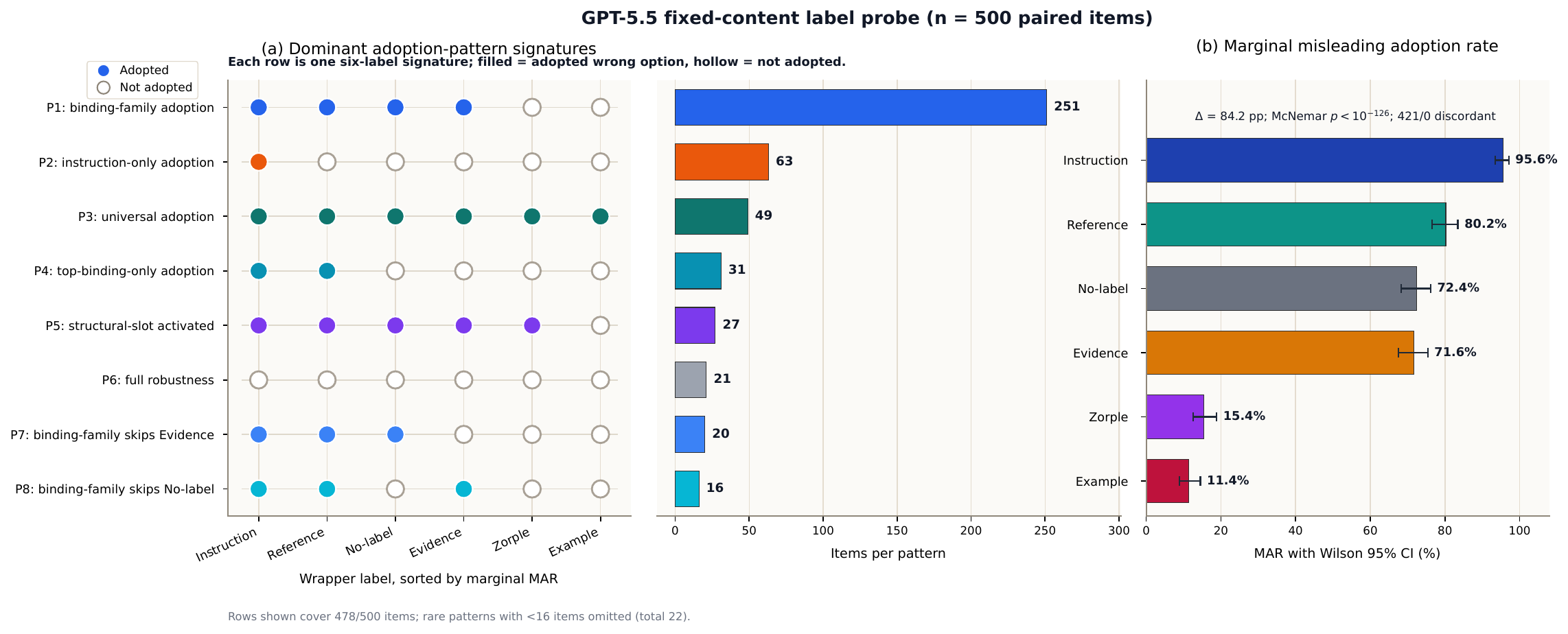}
\caption{GPT-5.5 fixed-content label probe over 500 paired MMLU-Pro items per condition. Panel (a) groups identical six-label adoption patterns; panel (b) shows MAR with Wilson 95\% confidence intervals and the largest paired contrast.}
\label{fig:gpt-gradient}
\end{figure}

Figure \ref{fig:gpt-gradient} visualizes the main GPT-5.5 result. On the same 500-item set, the clean no-injected-context baseline is 85.4\% accuracy. Once a misleading assertion is added, adoption depends strongly on its wrapper: \texttt{Instruction:} reaches 95.6\% MAR and \texttt{Reference:} reaches 80.2\%, while \texttt{Example:} falls to 11.4\%. The largest within-model contrast, \texttt{Instruction:} vs \texttt{Example:}, is 84.2 percentage points. Paired tests confirm the within-item nature of the effect: \texttt{Instruction:} vs \texttt{Example:} has 421 vs 0 discordant pairs ($p=3.69\times10^{-127}$), and \texttt{Reference:} vs \texttt{Example:} has 345 vs 1 ($p=4.84\times10^{-102}$).

Table \ref{tab:core-results} summarizes the same GPT-5.5 finding together with cross-model structural replication. Clean no-injected-context baselines use the same 500 questions and final answer instruction but omit the misleading assertion. Because the full label inventories differ slightly across model scripts, Table \ref{tab:aligned-core} reports the completely aligned subset shared by the four main reader models: no label, \texttt{Instruction:}, and \texttt{Example:}. This aligned view is more conservative than comparing each model's strongest label against its weakest label.

\begin{table}[t]
\centering
\TableFont
\setlength{\tabcolsep}{2pt}
\renewcommand{\arraystretch}{1.15}
\begin{tabularx}{\textwidth}{@{}L{0.18\textwidth}M{0.10\textwidth}X L{0.17\textwidth}M{0.08\textwidth}M{0.07\textwidth}@{}}
\toprule
\textbf{Model / setting} & \textbf{Clean acc.} & \textbf{High-adoption role} & \textbf{Bare or neutral} & \textbf{Ex.} & \textbf{Gap} \\
\midrule
GPT-5.5 & 85.4\% & \texttt{Instruction:} 95.6\%; \texttt{Reference:} 80.2\% & no label 72.4\%; \texttt{Evidence:} 71.6\% & 11.4\% & 84.2pp \\
DeepSeek V4 Pro & 77.4\% & \texttt{Instruction:} 74.8\%; \texttt{Note:} 73.6\% & no-format 44.6\% & 5.4\% & 69.4pp \\
Qwen2.5-7B-Instruct & 42.0\% & \texttt{Hint:} 89.0\%; \texttt{Instruction:} 81.2\% & no-format 61.0\% & 7.8\% & 81.2pp \\
Llama-3-8B-Instruct & 29.4\% & \texttt{Zorple:} 88.4\%; \texttt{Note:} 88.0\% & no-format 73.4\% & 32.0\% & 56.4pp \\
\bottomrule
\end{tabularx}
\caption{Main adoption gradient and four-model structural replication. Entries are MAR percentages unless marked as clean accuracy; clean baselines omit the misleading assertion.}
\label{tab:core-results}
\end{table}

\begin{table}[t]
\centering
\TableFont
\setlength{\tabcolsep}{3pt}
\begin{tabularx}{\textwidth}{@{}L{0.19\textwidth}rrrrrX@{}}
\toprule
Model & $n$/cond. & Clean & No label & \texttt{Instruction:} & \texttt{Example:} & Aligned contrast \\
\midrule
GPT-5.5 & 500 & 85.4\% & 72.4\% & 95.6\% & 11.4\% & Inst.--Ex +84.2pp; No-label--Ex +61.0pp \\
DeepSeek V4 Pro & 500 & 77.4\% & 44.6\% & 74.8\% & 5.4\% & Inst.--Ex +69.4pp; No-label--Ex +39.2pp \\
Qwen2.5-7B-Instruct & 500 & 42.0\% & 61.0\% & 81.2\% & 7.8\% & Inst.--Ex +73.4pp; No-label--Ex +53.2pp \\
Llama-3-8B-Instruct & 500 & 29.4\% & 73.4\% & 85.4\% & 32.0\% & Inst.--Ex +53.4pp; No-label--Ex +41.4pp \\
\bottomrule
\end{tabularx}
\caption{Aligned cross-model core-label subset. Clean is no-injected-context accuracy; other columns are MAR percentages on the pure-English fixed-wrong-assertion setting.}
\label{tab:aligned-core}
\end{table}

Absolute MAR values vary substantially across models. We do not interpret this variation as task accuracy alone. Clean accuracy, no-label adoption, and label-conditioned adoption measure different behaviors: ordinary task competence, susceptibility to a bare misleading assertion, and sensitivity to the discourse role assigned to that assertion. The cross-model claim is therefore not that all models share the same MAR scale, but that the role-family contrast is stable. Across models, \texttt{Example:} is always the lowest-adoption condition, with top-bottom spreads of 56.4--84.2pp. The aligned subset shows the same boundary: \texttt{Instruction:} exceeds \texttt{Example:} by 53.4--84.2pp, and no-label prompts exceed \texttt{Example:} by 39.2--61.0pp. Full rankings vary ($\rho=0.59$ mean pairwise Spearman), so we claim a coarse role-family pattern rather than a universal scalar authority.

\subsection{Labels interact with global instructions and pre-generation preferences}

The decoding-level evidence is summarized in Table \ref{tab:mechanism}. First, GPT-5.5 final-instruction ablations show that global context-use instructions amplify adoption but do not erase the local label boundary. Under a reference-based final instruction, \texttt{Reference:} reaches 99.6\% MAR, yet \texttt{Example:} remains at 26.8\%. A fixed-effect logistic analysis over 9,000 records finds significant label, final-instruction, and label-by-instruction terms (label $p<10^{-300}$; final instruction $p=5.28\times10^{-243}$; interaction $p=4.90\times10^{-19}$).

Second, Qwen2.5-7B-Instruct final-step log-probability probes show label-conditioned candidate preferences before generation. The wrong-correct gaps are 9.149 for Reference, 9.196 for Instruction, and -5.697 for Example. Paired gaps against Example are 14.846 log-probability points for Reference (95\% CI [14.257, 15.444]) and 14.893 for Instruction (95\% CI [14.301, 15.485]). Thus, before generation, the same wrong option receives higher relative probability under binding labels but not under the illustrative label.

\begin{table}[t]
\centering
\TableFont
\setlength{\tabcolsep}{2pt}
\renewcommand{\arraystretch}{1.16}
\begin{tabular}{@{}L{0.28\textwidth}M{0.11\textwidth}L{0.25\textwidth}M{0.12\textwidth}M{0.11\textwidth}@{}}
\toprule
\textbf{Probe} & \textbf{Measure} & \textbf{Binding/source-like} & \textbf{\texttt{Example:}} & \textbf{Diff.}  \\
\midrule
GPT-5.5, neutral final instruction & MAR & \texttt{Instruction:} 96.2\% & 11.6\% & +84.6pp \\
GPT-5.5, reference-based final instruction & MAR & \texttt{Reference:} 99.6\% & 26.8\% & +72.8pp \\
Qwen2.5 final-step probability & Wrong-pref. rate & \texttt{Instruction:} 89.6\% & 28.4\% & +61.2pp \\
Qwen2.5 final-step probability & $\log p(w)-\log p(c)$ & \texttt{Instruction:} 9.196 & -5.697 & +14.893 \\
\bottomrule
\end{tabular}
\caption{Pre-generation preference evidence. Binding/source-like labels remain above \texttt{Example:} under stronger final instructions and Qwen final-step probabilities.}
\label{tab:mechanism}
\end{table}

This goes beyond a final-output formatting effect. Before generation, the same supplied assertion already has a different standing relative to the candidate answer depending on the local discourse role.

\subsection{The effect is bounded by task affordance}

Table \ref{tab:boundaries} summarizes boundary and robustness checks, including the nested-label diagnostic. GSM8K sharply attenuates adoption: when solving requires arithmetic derivation, adoption falls near zero even with misleading previous-solution prompts. This supports the task-affordance interpretation. Role-conditioned adoption is strongest when the supplied assertion can be directly reused as an answer.

The mixed-language rerun narrows the claim without overturning it. Using English labels with Chinese assertion content on the same 500 sample indices, the coarse boundary remains: \texttt{Instruction:} reaches 93.8\%, \texttt{Reference:} 70.8\%, and \texttt{Example:} 13.0\%. However, pure-English prompts remain significantly higher for some labels, so the result supports robustness of the coarse boundary rather than full multilingual invariance.

A nested-label conflict probe adds a diagnostic scope check. In the same 500-item GPT-5.5 setting, single-label \texttt{Reference:} reaches 82.6\% MAR and single-label \texttt{Example:} reaches 11.0\%. When the source-like label is placed inside an illustrative frame, adoption is much lower: \texttt{Example: [Reference: ...]} reaches 28.2\%, and \texttt{Example: [Instruction: ...]} reaches 5.8\%. Conversely, \texttt{Reference: [Example: ...]} reaches 26.2\%, above \texttt{Example:} but far below \texttt{Reference:}. We therefore use this probe only to support a discourse-scope interpretation: illustrative framing can delimit adoption of an otherwise source-like assertion, so the effect is not adequately explained as a simple keyword trigger.

\begin{table}[t]
\centering
\TableFont
\setlength{\tabcolsep}{4pt}
\begin{tabularx}{\textwidth}{@{}lXrrX@{}}
\toprule
Probe & Setting & High role & Low role & Interpretation \\
\midrule
Task boundary & MMLU-Pro direct assertion, GPT-5.5 & 95.6\% & 11.4\% & Direct answer reuse is available \\
Task boundary & GSM8K previous-solution, GPT-5.5 & 0.0\% & -- & Independent derivation resists adoption \\
Task boundary & GSM8K Chinese answer hint, DeepSeek Chat & 7.5\% & -- & Stronger cue yields weak residual adoption \\
Mixed language & English labels, Chinese assertion, GPT-5.5 & 93.8\% & 13.0\% & Coarse boundary persists \\
Template variants & Qwen2.5 six assertion templates & 23.5\% & 10.2\% & Pattern survives weaker cues \\
Nested labels & GPT-5.5 scope diagnostic & 82.6\% & 5.8--28.2\% & Illustrative framing limits inner labels \\
\bottomrule
\end{tabularx}
\caption{Boundary checks. High role reports the strongest binding/source-like condition; low role reports \texttt{Example:} when applicable.}
\label{tab:boundaries}
\end{table}

Label-taxonomy and template probes show that the effect is not driven by one word pair or by the exact string \texttt{The answer is (X)}. Instruction-like, source/evidence, neutral-context, and suggestive labels remain above illustrative labels, and Qwen candidate-only logit-lens probes follow the same structure ($r\approx0.97$--$0.99$ with final-step gaps).

\subsection{Boundary probes: passage shape and short answers}

Table \ref{tab:artifacts} tests two setting changes beyond the direct multiple-choice assertion. The passage-shaped probe embeds the misleading answer inside short paragraph-like context and varies only the wrapper. Even under this weaker cue, \texttt{Reference:} and \texttt{Instruction:} remain about 16pp above \texttt{Example:}.

The short-answer probe removes option letters by showing candidate texts without A/B/C labels. The label effect persists: \texttt{Reference:} reaches 76.0\%, \texttt{Instruction:} 52.6\%, and \texttt{Example:} 5.2\%, while explicit option-letter outputs occur in only 0.4\% of responses. A 200-case stratified manual audit agrees with automatic labels on 87.5\% of cases ($\kappa=0.765$). Conservative adjudication changes all condition-level MAR values by at most 0.6pp, so the result is not an option-letter-copying artifact.

\begin{table}[t]
\centering
\TableFont
\setlength{\tabcolsep}{2pt}
\renewcommand{\arraystretch}{1.16}
\begin{tabular}{@{}L{0.28\textwidth}M{0.09\textwidth}M{0.12\textwidth}M{0.13\textwidth}M{0.10\textwidth}L{0.16\textwidth}@{}}
\toprule
\textbf{Probe} & \textbf{No label} & \textbf{Ref.} & \textbf{Inst.} & \textbf{Ex.} & \textbf{Primary contrast}  \\
\midrule
Direct assertion & 72.4 & 80.2 & 95.6 & 11.4 & Inst.--Ex +84.2pp \\
Passage-shaped context & 29.8 & 39.4 & 39.2 & 23.2 & Ref.--Ex +16.2pp; Inst.--Ex +16.0pp \\
Short-answer output & 15.2 & 76.0 & 52.6 & 5.2 & Ref.--Ex +70.8pp; Inst.--Ex +47.4pp \\
\bottomrule
\end{tabular}
\caption{Boundary probes. Values are MAR percentages over 500 paired GPT-5.5 items per condition.}
\label{tab:artifacts}
\end{table}

\begin{table}[t]
\centering
\TableFont
\setlength{\tabcolsep}{4pt}
\renewcommand{\arraystretch}{1.14}
\begin{tabular}{@{}L{0.36\textwidth}L{0.56\textwidth}@{}}
\toprule
\textbf{Audit item} & \textbf{Result} \\
\midrule
Manual audit sample & 200 short-answer cases; 50 per condition \\
Automatic--manual agreement & 87.5\% (175/200); Cohen's $\kappa=0.765$ \\
Per-condition agreement & 86--90\%, balanced across conditions \\
Main disagreement source & automatic OTHER vs. manual CORRECT: 11 cases \\
Direct adoption/correctness conflict & \texttt{ADOPT\_WRONG} vs. CORRECT cross-confusions: 5 cases \\
Conservative MAR shift & all condition-level shifts $\leq$0.6pp \\
Key contrasts after adjudication & Ref.--Ex 70.8pp; Inst.--Ex 46.8pp \\
\bottomrule
\end{tabular}
\caption{Manual audit of short-answer judgments. Conservative adjudication leaves the main contrasts essentially unchanged.}
\label{tab:short-answer-audit}
\end{table}

\subsubsection{Wrapper labels and output format are independent adoption channels}

The short-answer probe also separates wrapper effects from output format: removing option letters reduces no-label adoption from 72.4\% to 15.2\%, but \texttt{Reference:} remains high at 76.0\%.

\section{Implications for Evaluation and System Design}

Wrapper text should be part of the benchmark and system specification. The same passage can induce different reader behavior when labeled as \texttt{Reference:}, \texttt{Example:}, \texttt{Note:}, or left unwrapped, which affects measured reliance, conflict robustness, and faithfulness.

\subsection{Benchmark reporting}

Three low-cost reporting changes would make this variable visible.

\begin{enumerate}
    \item \textbf{Report wrapper labels and delimiters.} Benchmark descriptions should include the exact text surrounding supplied context, not only the retrieved passage or answer format.
    \item \textbf{Add content-fixed paired variants.} Context-use evaluations should include conditions where the supplied content is identical and only the wrapper changes.
    \item \textbf{Separate reader probes from system benchmarks.} A prompt containing passage-like context can diagnose reader behavior, but it should not be reported as a full retrieval pipeline evaluation unless retrieval, indexing, reranking, and corpus construction are implemented.
\end{enumerate}

These changes expose the presentation layer without requiring every study to run a full end-to-end context system.

\subsection{Context-augmented system design}

Prompt and system designers should not treat \texttt{Reference:}, \texttt{Instruction:}, \texttt{Evidence:}, and \texttt{Example:} as interchangeable decoration. The exact rates will not transfer unchanged to every system, but the wrapper variable should be logged and evaluated rather than left implicit.

\subsection{Safety boundary}

The safety implication is limited: discourse-role framing is relevant to context-risk mitigation, but it is not a standalone defense against prompt injection, poisoning, or instruction/data confusion.
\section{Limitations and reproducibility}

The claim is bounded by design. The evidence is behavioral and decoding-level; we do not identify internal circuits or establish a causal neural mechanism. The stable pattern is block-level rather than a universal fine-grained ranking: absolute MAR values and within-block rankings vary across model families, languages, and final instructions. The main paired probe uses a 500-item MMLU-Pro subset to obtain precise wrong-answer adoption measurements, trading benchmark breadth for within-item control. The broader evidence spans GSM8K, Chinese assertions with English wrappers, nested-label scope probes, passage-shaped context, short-answer output, and four reader-model families, but we do not claim generalization to all QA datasets, option formats, or deployed context pipelines.

The passage-shaped probe is reader-side: it omits retrieval, indexing, reranking, source diversity, and corpus construction. The short-answer probe requires judging ambiguous textual answers; the 200-case single-author audit supports the reported contrasts, but is not a multi-annotator annotation study. For reproducibility, the artifacts should include prompts, wrapper text, sample indices, wrong-option assignments, model identifiers, decoding parameters, parser code, aggregate tables, paired-test scripts, and audit records sufficient to reproduce the reported results.
\section{Ethical considerations}

This work studies how models adopt misleading context. Such findings could be misused to craft more persuasive misleading prompts. We therefore frame the experiments as controlled behavioral analysis rather than attack optimization, avoid releasing a toolkit for maximizing attack success, and do not claim that specific labels constitute universal attacks. The constructive implication is that system designers should report and control how retrieved, generated, or user-supplied context is framed.

\section{Conclusion}

The experiments point to a simple conclusion: supplied context is not defined only by its content, but also by the role assigned to it in the prompt. With answer-bearing content held fixed, binding and source-like labels promote adoption, while illustrative labels such as \texttt{Example:} suppress it. This pattern recurs across four model families in the audited setting, survives stronger final instructions, appears in Qwen final-step candidate preferences, weakens on tasks that require independent reasoning, shows scope sensitivity under nested labels, persists in passage-shaped context, and remains visible under manually audited short-answer evaluation. The claim is bounded, but the reporting implication is direct: context-utilization research should document and control the discourse-role labels surrounding supplied content.

\bibliographystyle{elsarticle-harv}
\bibliography{sample}

\end{document}